\documentclass[runningheads]{llncs}

\usepackage{amsmath}
\usepackage{graphicx}
\usepackage{bm}
\usepackage{adjustbox}
\usepackage{algorithm}
\usepackage{algorithmic}
\usepackage{enumitem}
\usepackage{amsthm}
\usepackage{xcolor}
\usepackage{tabularx}
\usepackage[stable]{footmisc}
\usepackage[caption=false]{subfig}
\usepackage{epsfig}
\usepackage{float}
\usepackage{adjustbox}
\newtheorem*{researchQuestion}{Research Question}
\definecolor{ForestGreen}{HTML}{228B22}

\newcommand{\citep}[1]{\cite{#1}}
\newcommand{\citet}[1]{\cite{#1}}

\newcommand{\EXCLUDE}[1]{}

\usepackage[T1]{fontenc}
\usepackage[acronym,shortcuts,smallcaps]{glossaries}

\newglossaryentry{LSTM}
{
  name={LSTM},
  description={long short-term memory},
  first={\glsentrydesc{LSTM} (\glsentrytext{LSTM})},
  plural={LSTMs},
  descriptionplural={long short-term memories},
  firstplural={\glsentrydescplural{LSTM} (\glsentryplural{LSTM})}
} 

\newglossaryentry{RNN}
{
  name={RNN},
  description={recurrent neural network},
  first={\glsentrydesc{RNN} (\glsentrytext{RNN})},
  plural={RNNs},
  descriptionplural={recurrent neural networks},
  firstplural={\glsentrydescplural{RNN} (\glsentryplural{RNN})}
} 

\newglossaryentry{NLP}
{
  name={NLP},
  description={natural language processing},
  first={\glsentrydesc{NLP} (\glsentrytext{NLP})},
} 

\newglossaryentry{MT}
{
  name={MT},
  description={machine translation},
  first={\glsentrydesc{MT} (\glsentrytext{MT})},
}

\newglossaryentry{ML}
{
  name={ML},
  description={machine learning},
  first={\glsentrydesc{ML} (\glsentrytext{ML})},
} 

\newglossaryentry{DL}
{
  name={DL},
  description={deep learning},
  first={\glsentrydesc{DL} (\glsentrytext{DL})},
} 

\newglossaryentry{HAN}
{
  name={HAN},
  description={hierarchical attention network},
  plural={HANs},
  descriptionplural={hierarchical attention networks},
  first={\glsentrydesc{HAN} (\glsentrytext{HAN})},
  firstplural={hierarchical attention networks (\glsentryplural{HAN})}
} 

\newglossaryentry{HAN-ST}
{
  name={HAN$_{\textrm{ST}}$},
  description={hierarchical attention network with structure tags},
  descriptionplural={hierarchical attention networks with structure tags},
  first={\glsentrydesc{HAN-ST} (\glsentrytext{HAN-ST})},
  firstplural={hierarchical attention networks with structure tags (\glsentryplural{HAN})}
}

\usepackage{mathtools}
\DeclarePairedDelimiter{\ceil}{\lceil}{\rceil}

\usepackage[urlcolor=blue]{hyperref}

\begin{document}
%

\title{ Active learning for reducing labeling effort in text classification tasks 
}

\author{Pieter Floris Jacobs\inst{1}\orcidID{0000-0002-8835-6356} \and \\ Gideon Maillette de Buy Wenniger \inst{2,1}\orcidID{0000-0001-8427-7055} \and \\
Marco Wiering \inst{1}\orcidID{0000-0003-4331-7537}
\and \\ Lambert Schomaker \inst{1}\orcidID{0000-0003-2351-930X}}

\authorrunning{P.F. Jacobs et al.}
%
\institute{University of Groningen, Groningen, The Netherlands \\
\email{p.f.jacobs AT student.rug.nl; \\\{l.r.b.schomaker, m.a.wiering \} AT rug.nl } 
\and
Open University of the Netherlands \\
\email{gemdbw AT gmail.com}\\
}
%
\renewcommand{\lastandname}{\unskip,}
\maketitle              
\begin{abstract}
  Labeling data can be an expensive task as it is usually performed manually by domain experts. This is cumbersome for deep learning, as it is dependent on large labeled datasets. Active learning (AL) is a paradigm that aims to reduce labeling effort by only using the data which the used model deems most informative. Little research has been done on AL in a text classification setting and next to none has involved the more recent, state-of-the-art Natural Language Processing (NLP) models. Here, we present an empirical study that compares different uncertainty-based algorithms with BERT$_{base}$ as the used classifier. We evaluate the algorithms on two NLP classification datasets: Stanford Sentiment Treebank and KvK-Frontpages. Additionally, we explore heuristics that aim to solve presupposed problems of uncertainty-based AL; namely, that it is unscalable and that it is prone to selecting outliers. Furthermore, we explore the influence of the query-pool size on the performance of AL. Whereas it was found that the proposed heuristics for AL did not improve performance of AL; our results show that using uncertainty-based AL with BERT$_{base}$ outperforms random sampling of data. This difference in performance can decrease as the query-pool size gets larger.

\end{abstract}

\section{Introduction}\label{sec:introduction}
 Deep Learning \glsunset{DL}(\Ac{DL}) is a field in machine learning in which neural networks with a large number of layers are made to perform complicated human tasks. These networks have to be trained on a large amount of data to be able to learn the underlying distribution of the task they are trying to model. In supervised learning, this data is required to be labeled with the desired output. This allows the network to learn to map the input to the desired output. This study will focus on an instance of supervised learning, called text classification. 
Data labeling is usually done manually and can grow to be an expensive and time-consuming task for larger datasets, like those used in \ac{DL}. This begs the question of whether there is no way to reduce the labeling effort while preserving good performance on the chosen task. Similarly to lossy compression \citep{Ahmed74}, we want to retain a good approximation of the original dataset while at the same time reducing its size as much as possible. More specifically: given a training set, how can we optimally choose a limited number of examples based on the amount of relevant information they contain for the target task? 

Conceptually, answering this question requires quantifying the amount of information contained in each data point. This finds its roots, like lossy compression, in information theory \citet{Shannon48}. 
A model trained on limited data has an entropy associated with its target variable predictions. Our goal is to greedily select the data for labeling, 
while reducing entropy as much as possible, similar to how it is done in research on decision trees \citep{Gulati2016}. In essence, we aim to 
incrementally, optimally select a subset of data points; such that the distribution encoded by the learned model maximizes the information gain or equivalently minimizes the Kullback-Leibler divergence \citep{Kullback51} with respect to the unknown distribution of the full labeled data. 
However, there are two problems. First, the labels of the data are not known until labeling, and additional held-out labeled data to aid the selection is typically not available either. This contrasts with the easier case of summarizing a known dataset by a subset of data, in which the Kullback-Leibler divergence of a selected subset with
the full set can be measured and minimized. Second, because the parameters of a neural network change during training, predictions and certainty of new data points also change. Because of these two problems, examples can only be greedily selected based on their expected utility for improving the current, incrementally improved model. As the actual labels for examples are lacking before their selection, their real utility cannot be known during selection. Therefore, only proxies for this utility such 
as model uncertainty can be used, as discussed next.

A machine-learning technique called Active Learning (AL) \citep{Settles2010} can be used to combat these problems. In AL, a human labeler is queried for data points that the network finds most informative given its current parameter configuration. The human labeler assigns labels to these queried data points and then the network is retrained on them. This process is repeated until the model shows robust performance, which indicates that the data that was labeled is a sufficient approximation of the complete dataset. There are multiple types of informativeness by which to determine what data to query the oracle for. For instance calculating what results in the largest model change \citep{Cai2013} or through treating the model as a multi-arm bandit \citep{Bouneffouf2014}. However, the existing literature predominantly utilizes different measures of model uncertainty \citep{drost2020,Gal2016Thesis,Gal2016Dropout,GalIslamGhahramani2017,teye2018}, which is also done in this research. Bayesian probability theory provides us with the necessary mathematical tools to reason about uncertainty, but for \ac{DL} has its complications.
The reason is that (typical) neural networks, as used for classification and regression, are discriminative models. These produce a single output, a so called point estimate. Even in the case of softmax outputs this is not a true probability density function \cite{Gal2016Thesis,Gal2016Dropout}. Another view on this is that modern neural networks often lack adequate \emph{confidence calibration}, meaning they fail at predicting probability estimates representative of the
true correctness likelihood \cite{guo2017calibration}.

This poses a problem to Bayesian probability theory as it prevents us from being able to perform Bayesian inference. With Bayesian inference we can determine the probability of a certain output y* given a certain input point x*: 
\begin{equation}
    \label{eq:inference}
    p(y*|x*, X, Y) = \int{{p(y*|x*,\omega)p(\omega, X, Y)d\omega}}
\end{equation}
Unfortunately, for the discriminative neural network models there is no probability distribution: the output is always the same for a given input. What is more, even if we suppose the network was generative (Eq. \ref{eq:inference}), the integral is not analytically solvable due to the fact that we need to integrate over all possible parameter settings $\omega$.
However, it can be approximated. Existing literature has explored different methods of achieving this, with Monte Carlo Dropout (MCDO) being the most popular one \citep{drost2020,Gal2016Dropout,Tsymbalov2018}. In MCDO, the network applies dropout \citep{hinton2014} to make the network generative. Multiple stochastic forward passes are performed to produce multiple outputs for the same input. The outputs can then be used to summarize the uncertainty of the model in a variety of ways.

This research uses the MCDO approximation to compare different uncertainty-related AL query methods for text classification, noting there is still little literature on the usability of AL for modern NLP models.
We strive to answer the following research question: 
\begin{researchQuestion}
How can uncertainty-based Active Learning be used to reduce labeling effort for text classification tasks?
\end{researchQuestion} 

Where previous literature focused on comparing AL strategies on small datasets and on the test accuracy of the final classifier, this paper will try and explore the usability of AL on a real-world setting, in which factors like the effect of transfer learning and considerations such as scalability have to be taken into account. The goal is to reach a performance similar to the state-of-the-art text-classification models that use a large randomly sampled set of labeled examples as training set. This should show whether AL can be applied to reduce labeling effort.

\section{Related Work}
\noindent \textbf{Active Learning applied to Deep Learning for Image Classification} \\
Multiple methods of incorporating AL into Deep Neural Networks (DNNs) have been proposed in the past. Most of these focus on image classification tasks.

Houlsby et al. \cite{houlsby2011bayesian} proposed an information theoretic approach to AL: Bayesian Active Learning by Disagreement (BALD). In hopes of achieving state-of-the-art performance and making minimal approximations for achieving tractability, they used a Gaussian process classifier and compared the performance of BALD to nine other AL algorithms. Their findings included that BALD, which we use in this study, makes the smallest number of approximations across all tested algorithms.

Gal et. al \cite{GalIslamGhahramani2017} used a Bayesian convolutional network together with MCD to be able to approximate Bayesian inference and thereby proposed an AL framework that makes working with high dimensional data possible. They compared results of a variety of uncertainty-based query functions (including BALD and variation ratio) to random sampling and found that their approach to scaling AL to be able to use high dimensional data was a significant improvement to previous research, with variation ratio achieving the best results.

Drost \cite{drost2020} provided a more extensive discussion of the different ways of incorporating uncertainty into DNNs. He tried to learn which way of computing the uncertainty for DNNs worked best. Using a convolutional neural network, he compared the use of dropout, batch normalization, using an Ensemble of NNs and a novel method named Error Output for approximating Bayesian inference. His main conclusion was that using dropout, batch normalization and ensembles were all useful ways of lowering uncertainty in model predictions. He found that the Ensemble method provided the best uncertainty estimation and accuracy but that it was very slow to train and required a large amount of memory. He concluded MCDO, which is what we use in this study, to be a promising strategy of uncertainty estimation, albeit that one has to take into account slow inference times. 

Gikunda and Jouandeau \cite{budgetActiveLearning2021} explored an approach for preventing the selection of outlier examples. They combined the uncertainty measure with a correlation measure, measuring the correlation of each unlabeled example with all other unlabeled examples. A higher correlation indicated that an example was less likely to be an outlier. 
Their method is similar to using a local KNN-based example density as discussed in \cite{zhu-etal-2008-active}, which is one of the methods we used in this work. 
The main difference with the KNN-density approach is that their correlation-based density does not consider local neighborhoods in the density estimation. 
As uncertainty measure they used so-called sampling margin, which is based on the difference in probability between the most likely and second most likely class according to softmax outputs. This is somewhat similar to variation ratio, but does not use stochastic forward passes. It uses plain softmax outputs instead, making it quite distinct from the dropout-sampling based approach we adopt in this work. \\

\noindent \textbf{Active Learning applied to Deep Learning for Text Classification} \\
A survey of deep learning work on using AL for text classification is given in \cite{surveyActiveLearningForDeepTextClassification2020}. They present a taxonomy of different query functions, including those focused on prediction and model uncertainty that we use. They also discuss the incorporation of word embeddings into DNN-based AL, which is something that we attempt in this study.

BERT is used in combination with AL in \cite{ein-dor-etal}.
They presented a large-scale empirical study on AL techniques for BERT-based classification, covering a diverse set of AL strategies and datasets; focusing on binary text classification with small annotation budgets. They concluded that AL can be used to boost BERT performance.\\

\noindent \textbf{Active Learning for Regression} \\
Whereas our work is on classification, dropout-based AL can be adapted for regression as well, and this was done by \cite{dropoutBasedActiveLEarningForRegression2018}. They used the set of $T$ sample predictions from the forward passes to compute sample standard deviation for the $T$ predictions, using this as a measure of uncertainty. Evaluation was done on standard open multivariate datasets of the UCI Machine Learning repository. \\
 
\noindent \textbf{Confidence Calibration} \\
Dropout sampling as used in this work aims to solve the problem that softmax outputs are not reliable representations of the true class probabilities. This problem is known as \emph{confidence calibration}, and dropout sampling is not the only solution to it. 

Guo et. al \cite{guo2017calibration} evaluated the performance of various post-processing techniques that took the neural network outputs and transformed them into values closer to representative probabilities. They found that in particular a simplified form of \emph{Platt Scaling}, known as \emph{temperature scaling}, was effective in calibrating predictions on many datasets. This method conceptually puts a logistic regression model with just one learnable 'temperature' parameter behind the softmax outputs, and is trained by optimizing negative log likelihood (NLL) loss over the validation set. It thus learns to spread out or peak the probabilities further in a way that helps to decrease NLL loss, thereby as a side-effect increasing calibration.  
Recently, using a new procedure inspired by Platt Scaling, Kuleshov et. al \cite{pmlr-v80-kuleshov18a} generalized an effective approach for confidence calibration to be usable for regression problems as well.\\

\section{Methods}\label{sec:methods}
This section will go on to describe the general AL loop, the model architecture, the used query functions, the implemented heuristics, and finally the experimental setup.

\subsection{Active Learning}
An implementation of the general AL loop/round is shown in Appendix \ref{ap:appendix_algorithm} (Algorithm \ref{alg:al-loop}). It consists of four steps:
    \begin{enumerate}
        \item \textbf{Train:} The model is reset to its initial parameters. After this, the model is trained on the labeled dataset $\mathcal{L}$. The model is reset before training because otherwise the model would overfit on data from previous rounds \citep{PartialFeedBackHu2018}. 
        \item \textbf{Query:} A predefined query function is used to determine what data is to be labeled in this AL round. As discussed, this can be done in various ways, but the guiding principle is that the data that the model finds most useful for the chosen task gets queried.
        \item \textbf{Annotate:} The queried data is parsed to a human expert, often referred to as the oracle. The oracle then labels the queried examples.
        \item \textbf{Append:} The newly-labeled examples are transferred from the unlabeled dataset $\mathcal{U}$ to $\mathcal{L}$. The model is now ready to be retrained to recompute the informativeness of the examples in $\mathcal{U}$ now that the underlying distribution of $\mathcal{L}$ has been altered.

    \end{enumerate}
Please note that the datasets used for the experiments (Section \ref{ssec:Data}) were fully labeled and the annotation step thus got skipped in this research. $\mathcal{U}$ existed out of labeled data that was only trained on from the moment it got queried. This was done to speed up the process and to enable scalable and replicable experiments with varying experimental setups.

\subsection{Model Architecture}
\subsubsection{BERT}
The model used to classify the texts was BERT$_{base}$ \citep{devlin2019bert}, a state-of-the-art language model which is a variant of the Transformer model \citep{vaswani2017attention}. 
Specifically, we used the uncased version of BERT$_{base}$, as the information of capitalization and accent markers was judged to be not helpful for the used tasks and datasets. Due to computational constraints, only the first sentence of the used texts was put into the tokenizer and the maximal length to which the tokenizer either padded or cut down this sentence was set to 50.
To better deal with unknown words and shorter text, we used the option of the BERT$_{base}$ tokenizer to make use of special tokens for sentence separation, padding, masking and to generalize unknown vocabulary. 
Finally, a softmax layer was added to the end of BERT$_{base}$, which is essential as the implemented query functions (Section \ref{ssec:query_function}) compute uncertainty based on sampled output probability distributions.

\subsubsection{Monte Carlo Dropout}\label{ssec:MCD}
Monte Carlo dropout (MCDO) is, as discussed in Section \ref{sec:introduction}, a technique that enables reasoning about uncertainty with neural networks. Dropout \citep{hinton2014} essentially 'turns off' neurons during the forward pass with a predefined probability. Dropout is normally used during training to prevent overfitting and create a more generalized model. In MCDO though, it is used to approximate Bayesian inference \citep{Gal2016Dropout} through creating $T$ predictions for 
all data points, using $T$ slightly different models induced by different dropout samples. 
The result of these so-called stochastic forward passes (SFP's) can then be used by the query function to compute the uncertainty, as will be explained in Section \ref{ssec:query_function}. The way MCDO is incorporated in the AL loop is shown in green in the Appendix (Algorithm \ref{alg:al-loop-MCD}). 
BERT$_{base}$ has two different types of dropout layers: hidden dropout and attention dropout. 
Both were turned on when performing a stochastic forward pass. 
Note that there are other ways of approximating Bayesian inference with neural networks. Frequently used ones are:
\begin{itemize}
    \item Having an ensemble of neural networks vote on the label \citep{ensembles1994}.
    \item Monte Carlo Batch Normalization (MCBN) \citep{teye2018}.
\end{itemize}
MCDO was chosen over the ensemble method due to it being easier to implement and quicker to train. MCBN was not chosen as it has been shown to be more inconsistent than MCDO \citep{drost2020}. 

\subsubsection{Sentence-BERT}\label{ssec:SentenceBERT}
Textual data offers the advantage of having access to the use of pre-trained word embeddings. These are learned representations of words into a vector space in which semantically similar words are close together. 
Textual embeddings can be computed in a variety of ways. BERT specific ones include averaging the pooled BERT embeddings and looking at the BERT CLS token output. Other more general ways are averaging over Glove word embeddings \citep{pennington-etal-2014-glove} and averaging embeddings created by a Word2Vec model \citep{Mikolov2013EfficientEO}. We have opted to make use of Sentence-BERT \citep{Reimers2019}, a Siamese BERT architecture trained to produce embeddings that can be adequately compared using cosine-similarity. For our purposes this provides better performance than the other embedding computations. Sentence-BERT was used separately from the previously discussed BERT$_{base}$ model, and was used only for assigning embeddings to each sentence in the dataset that were used by the heuristics described in Section \ref{sssec:RE}.

\subsection{Query Functions}\label{ssec:query_function}
The query functions determine data selection choices of the model in the AL loop. This paper will focus on functions that reason about uncertainty, obtained from approximated Bayesian distributions \citep{Gal2016Dropout}. For every data point, the distribution is derived from $T$ stochastic forward passes and resulting $T$ (in our case) softmax probability distributions. The following subsections will go on to discuss the implemented query functions. One is encouraged to look at \citet{Gal2016Thesis} for an extensive discussion that highlights the difference between these functions. 

\subsubsection{Variation Ratio}\label{sssec:variation_ratio}
The variation ratio is a measure of dispersion around the class that the model predicts most often (the mode). The intuition here is that the model is uncertain about a data point when it has predicted the mode class a relatively small number of times. This indicates that it has predicted other classes a relatively large number of times. Equation \ref{eq:variation_ratio} shows how the variation ratio is computed, where $f_x$ denotes the mode count and $T$ the number of stochastic forward passes. 
\begin{equation}
    v[x] = 1 - \frac{f_x}{T}
    \label{eq:variation_ratio}
\end{equation}
The function attains its maximum value when the model predicts all classes an equal amount of times and its minimum value when the model only predicts one class across all stochastic forward passes. Variation ratio only captures the uncertainty contained in the predictions, not the model, as it only takes into account the spread around the most predicted class. It is thus a form of predictive uncertainty.

\subsubsection{Predictive Entropy}\label{sssec:pred_entropy}
Entropy $H(x)$ in the context of information theory is defined as:
\begin{equation}
\label{eq:entropy}
    H(x) = -\sum_{i=1}^n p(x_i) \log_2 p(x_i)
\end{equation}
This formula expresses the entropy in bits per symbol to be communicated, in which $p(x_i)$ gives the probability of the $i$-th possible value for the symbol.
Entropy is used to quantify the information of data. In our case we want to know the chance of the model classifying a data point as a certain class given the input and model parameters ($p(y=c|\textbf{x}, \boldsymbol{\omega})$). We can compute this chance by averaging over the softmax probability distributions across the $T$ stochastic forward passes. This adjusted version of entropy is denoted in Equation \ref{eq:pred_entropy}, where $\hat{\omega_t}$ denotes the stochastic forward pass $t$, and $c$ the number associated to the class-label. 

\begin{equation}
    \begin{split}
    H[y|\textbf{x}, \mathcal{D}_{train}] = -\sum_c\left(\frac{1}{T}\sum_t p(y=c|\textbf{x},\boldsymbol{\hat{\omega}_t})\right)\noindent\\
    \log\left(\frac{1}{T}\sum_t p(y=c|\textbf{x}, \boldsymbol{\hat{\omega}_t})\right)
    \label{eq:pred_entropy}
    \end{split}
\end{equation}

To exemplify: in binary classification, the predictive entropy is highest when the model its softmax classifications consist of $T$ times  [0.5, 0.5]. In that case, expected surprise when we would come to know the real class-label is at its highest. The uncertainty is computed by averaging over all predictions and thus falls under predictive uncertainty.

\subsubsection{Bayesian Active Learning by Disagreement}\label{sssec:BALD}
Predictive entropy (Section \ref{sssec:pred_entropy}) is used to quantify the information in one variable. Mutual information or joint entropy is very similar but is used to calculate the amount of information one variable conveys about another. In our case, we'll be looking at what the average model prediction will convey about the model posterior, given the training data. This is a form of conditional mutual information, the condition or the third variable being the training data $\mathcal{D}_{train}$. Houlsby et al. \citet{houlsby2011bayesian} used this form of mutual information in an AL setting and dubbed it Bayesian active
learning by disagreement (BALD). 

\begin{equation}
    \begin{split}
    I[y, \omega|\textbf{x}, \mathcal{D}_{train}] = -\sum_c\left(\frac{1}{T}\sum_t p(y=c|\textbf{x},\boldsymbol{\hat{\omega}_t})\right)\\
    \log\left(\frac{1}{T}\sum_t p(y=c|\textbf{x}, \boldsymbol{\hat{\omega}_t})\right)\\
    -\frac{1}{T}\sum_{c,t}p(y=c|\textbf{x},\boldsymbol{\hat{\omega_t}})\\
    \log p(y=c|\textbf{x},\boldsymbol{\hat{\omega_t}})
    \label{eq:mutual_information}
    \end{split}
\end{equation}

The difference between Equations \ref{eq:mutual_information} and \ref{eq:pred_entropy} is that the conditional entropy is subtracted from the predictive entropy. The conditional entropy is the probability of the full output being generated from the training data and the input. This is the reason we do not average the predictions for every single class. We first sum over all classes, so that we do not average over the model parameters for every single class and thus take into account the fact that we are looking at the chance of the complete probability distribution being generated.

BALD is maximized when the $T$ predictions are strongly disagreeing about what label to assign to the example. So in the binary case, it would be highest when the predictions would alter between [1,0] and [0,1] as these two predictions are each others complete opposite. Unlike the variation ratio and predictive entropy,  BALD is a form of model uncertainty. When the softmax outputs would be equal to $T$ times [0.5,0.5], the minimal BALD value would be returned as the predictions are the same and the model is thus very confident about its prediction.

\subsection{Heuristics}\label{ssec:heuristics}
\subsubsection{Redundancy Elimination}\label{sssec:RE}
In AL, a larger query-pool size (from now on referred to as $q$) results in the model being retrained less and the uncertainties of examples being re-evaluated less frequently. Consequently, the model gets to make less informed decisions as it uses less up-to-date uncertainty estimates. Larger $q$ could therefore theoretically cause the model to collect many similar examples 
for specific example types with high model uncertainty in an AL round. 
Say for instance we were dealing with texts about different movie genres. Suppose the data contained a lot of texts about the exact same movie. When the model would be uncertain about this type of text, a large $q$ would result in a large amount of these texts getting queried. 
This could be wasteful, as querying this type of text a small amount of times would likely result in the model no longer being uncertain about that type of text.
Note however, that low model uncertainty by itself is no guarantee for robustly making accurate predictions for a type of examples. Yet provided such robust performance is achieved, additional examples of the same type would be a waste.

The above could form a problem as although a smaller $q$ should theoretically provide us with better results, it also requires more frequent uncertainties re-computation. Every computation of the uncertainties requires $T$ stochastic forward passes on the unlabeled dataset $\mathcal{U}$. This entails that, next to the computation, the time required to label a dataset would increase as well, which is not in line with our goal. In hopes of improving performance with larger $q$, we propose two heuristics:

\begin{enumerate}
    \item Redundancy Elimination by Training (RET)
    \item Redundancy Elimination by Cosine Similarity (RECS)
\end{enumerate}

For both of these heuristics, a new pool, which we will refer to as the redundancy-pool $\mathcal{RP}$, is introduced. The query-pool $\mathcal{QP}$ will be a subset of $\mathcal{RP}$ of which we will try to select the most dissimilar examples.

RET tries to eliminate redundant data out of $\mathcal{RP}$ by using it as a pool to retrain on. The data point with the highest uncertainty is trained on for one epoch and then the uncertainties of the examples in $\mathcal{RP}$ are recomputed. This process gets repeated until $\mathcal{QP}$ is of the desired size. Note that although this strategy seems similar to having a $q$ of one, it is less computationally expensive as only the uncertainties for the examples in $\mathcal{RP}$ have to be recomputed (which also shrinks after each repetition). Algorithm \ref{alg:AL-RET} of Appendix \ref{ap:appendix_algorithm} shows how RET is integrated in the AL loop. 

The main purpose of RET is to enable the use of larger $q$. However, one needs to be mindful of the fact that when $q$ is increased, $\mathcal{RP}$ is to be increased in size well. This being due to the fact that smaller differences between the sizes of $\mathcal{RP}$ and $\mathcal{QP}$ result in less influence of the heuristic. 
In the RET algorithm,  forward passes over $\mathcal{RP}$ contribute to the total amount of forward passes. Furthermore, this contribution increases linearly with the redundancy-pool size ($|\mathcal{RP}|$)  and in practice coupled query-pool size $q$. Using $|\mathcal{RP}|= 1.5 \times q$, 
this contribution starts to dominate the total amount of forward passes (approximately) once $q  > \sqrt{|\textrm{data}|}$.
This is explained in more detail in Appendix \ref{appendix:RET-algorithm-cost-analysis}.
This limits its use for decreasing computation by increasing $q$.
Because of this, RECS is aimed at being computationally cheaper. 

Instead of retraining the model and constantly taking into account recomputed uncertainties, RECS makes use of the sentence embeddings created by Sentence-BERT (Section \ref{ssec:SentenceBERT}). The assumption made is that semantically similar data conveys the same type of information to the model. The examples are selected based on their cosine similarity to other examples. $\mathcal{RP}$ is looped through and examples are only added to $\mathcal{QP}$ if their cosine similarity to all other points that are already in $\mathcal{QP}$ is lower than the chosen threshold $l$. If not enough examples are selected to get the desired $q$, the threshold gets decreased by 0.01. Algorithm \ref{alg:RECT} of Appendix \ref{ap:appendix_algorithm} shows how this heuristic is added to the AL loop. 

\subsubsection{Sampling by Uncertainty and Density (SUD)}\label{ssseq:SUD} Schomaker and Oosten \citet{Schomaker2013} showed that the distinction between separability and prototypicality is important to account for. In their use case of the SVM, data points that had a high margin to the decision boundary were not always representative of the class prototype. Uncertainty sampling also tries to sample examples close to the decision boundary, but has been shown to often select outliers \cite{Mccallum2001,Tang02activelearning}. 
Outliers contain a lot of information that the model has not encountered yet, but this information is not necessarily useful. As with the previously described RECS heuristic, we hypothesize that semantically similar sentences provide the same type of information. In that situation, outliers are very far from other examples in embedding space. 

Zhu et. al \citet{zhu-etal-2008-active} proposed a K-Nearest-Neighbor-based density approach called Sampling by Uncertainty and Density (SUD) to avoid outliers based on their distance in embedding space. In this approach, the mean cosine similarity between every data point and its $K$ most similar neighbors is computed. A low value indicates that a data point is not very similar to others. This value is then multiplied with the uncertainty and the dataset is sorted based on this Uncertainty-Density measure. They showed that this measure improved performance of the maximum entropy model classifier. We will explore whether this approach also works for BERT combined with the embeddings computed by Sentence-BERT. The adjusted pseudocode is shown in Appendix \ref{ap:appendix_algorithm} (Algorithm \ref{alg:al-SUD}).

\subsection{Experimental Setup\footnote{The code used for the experiments can be found at \url{https://github.com/Pieter-Jacobs/bachelor-thesis}}}
\subsubsection{Data}\label{ssec:Data}
Two datasets were used to validate and compare the performance of the different AL implementations. Table \ref{tab:datasets} shows an overview of the amount of examples and classes of each dataset. 
\begin{table}[htpb]
        \caption{An overview of the two datasets used in the experiments}
        \begin{center}
            \begin{tabularx}{\linewidth}{XXX}
                 Dataset & Examples & Number of Classes  \\
                 \hline
                 SST & 11,850 & 5 \\
                 KvK & 2212 & 15 \\
                 \hline
            \end{tabularx}
        \end{center}
        \label{tab:datasets}
\end{table}
The first of the used datasets was the Stanford Sentiment Treebank \citep{socher-etal-2013-recursive} (SST). SST exists out of 215,154 phrases from movies with fine-grained sentiment labels in the range of 0 to 1. These phrases are contained in the parse trees of 11,855 sentences. Only these full sentences were used in the experiments, and the sentiment labels were mapped to five categories in the following way:
 \begin{multicols}{2}
\begin{itemize}
    \item 0 $\leq$ label $<$ 0.2: very negative
    \item 0.2 $\leq$ label $<$ 0.4: negative
    \item 0.4 $\leq$ label $\leq$ 0.6: neutral
    \item 0.6 $<$ label $\leq$ 0.8: positive
    \item 0.8 $<$ label $\leq$ 1: very positive
    \item[] ~ 
\end{itemize}
\end{multicols}
Use of the SST dataset was motivated by its size as well as by it being a benchmark for language models. It allowed for the evaluation of AL for a larger dataset and for comparison with results found in related work such as \citet{munikar2019finegrained}. This helped to check whether BERT$_{base}$ was achieving desirable performance.

The second dataset that was used consists of the descriptions of companies located in Utrecht. The companies are all registered at the Dutch Chamber of Commerce, or Kamer van Koophandel (KvK) and were mapped to their corresponding SBI-code. The SBI code denotes the sector a company operates in, as defined by the KvK. The HTML of the companies websites was scraped and the meta content that was tagged as the description was extracted. In nearly all cases, this contained a short description about what the company was involved in. Note that only English descriptions were used. The KvK dataset provided us with the opportunity to evaluate AL for a classification problem with a large amount of classes as well as the ability to compare results between a dataset with a limited number of examples and one with a relatively large amount of examples (SST). Testing AL on a dataset with a limited number of examples was deemed necessary due to the fact that most of the positive results found in related work were achieved by making use of very small datasets. The dataset will not be shared and is not available online due to the fact that it was constructed as part of an internship at Dialogic. 

\subsubsection{Evaluation Metrics}
To evaluate and compare the performance of the different AL strategies, two evaluation metrics were reported: the accuracy and an altered version of the deficiency metric proposed in \citet{zhu-etal-2008-active}.

The variant of deficiency that was used is shown in Eq. \ref{eq:deficiency}, in which $n$ denotes the amount of accuracy scores, $acc(R)$ denotes the accuracy of the reference strategy and $acc(C)$ the accuracy of the strategy to be compared to this reference strategy. In our case, $n$ is equal to $\frac{|\mathcal{U}|}{q} + 1$ (+1 comes from the accuracy achieved after training on the seed), as we computed the test accuracy after every AL round.\footnote{For our experiments, this resulted in our $n$ ranging from 20 to 191 for the SST dataset and from 17 to 152 for the KvK dataset (the used $q$ can be found in Section \ref{sssec:experiments}).} Furthermore, instead of using the accuracy that was achieved in the final AL round for $acc(C)$ and $acc(R)$ like \citet{zhu-etal-2008-active}, we use the overall maximum accuracy. This accounts for the fact that the last achieved accuracy in a classification task is not necessarily the best value, while still returning a metric which provides a summary of the entire learning curve. This in turn means that a decrease/increase in its value is analogical to a decrease/increase in overall performance of the comparison strategy. However, the deficiency does not convey whether there were points at which the accuracy of a strategy was higher than usual and would serve as a good point to cut-down the dataset to reduce labeling effort. A deficiency of $<$1 indicates a better performance than the reference strategy whereas a value of $>$1 indicates a worse performance. 
\begin{equation}
    \label{eq:deficiency}
    DEF(AL, R) = \frac{\sum_{t=1}^n(max(acc(R))-acc_t(C)}{\sum_{t=1}^n(max(acc(R))-acc_t(R))}
\end{equation}
    
\subsubsection{Experiments}\label{sssec:experiments}
The goal of the experiments was to answer the question of whether overall labeling effort could be reduced through making use of AL. We split this into the following three sub-questions:
\begin{enumerate}
    \item Does AL achieve better performance with less data when compared to plain random sampling?
    \item What is the relation between query-pool size $q$ and the achieved performance?
    \item Do the proposed heuristics (SUD, RET, RECS) improve the performance of AL?
\end{enumerate}

The statistical setup used for the experiments can be found in Table \ref{tab:stat_setup}. The setup for SST was based on the proposed setup in \citet{socher-etal-2013-recursive}.
To reiterate, the following AL strategies were implemented:
 \begin{multicols}{2} 
\begin{enumerate}
    \item Variation Ratio (Section \ref{sssec:variation_ratio})
    \item Predictive Entropy (Section \ref{sssec:pred_entropy})
    \item BALD (Section \ref{sssec:BALD})
    \item RET (Section \ref{sssec:RE})
    \item RECS (Section \ref{sssec:RE})
    \item SUD (Section \ref{ssseq:SUD})
\end{enumerate}
\end{multicols}

\begin{table}[t!]
    \caption{The statistical setup used for both datasets. The percentages used are relative to the full dataset size.}
        \begin{tabularx}{\linewidth}{XXXXX}
                Dataset & Seed & $\mathcal{U}$ & Dev & Test\\
                \hline
                SST & 594 (5\%) & 7951 (67\%) & 1101 (9\%) & 2210 (19\%)\\
                KvK & 111 (5\%) & 1659 (75\%) & 221 (10\%) & 221 (10\%) \\
                \hline
        \end{tabularx}
    \label{tab:stat_setup}
\end{table}

To answer subquestion 1, these strategies were compared to the performance of random sampling using a $q$ of 1\% of the dataset size. For subquestion 2, the three query functions were be compared across three $q$: 0.5\%, 1\% and 5\% of the dataset size. Finally, to be able to answer subquestion 3, RET, RECT and SUD were compared with a $q$ of 1\%. As RET, RECS and SUD were meant as additions to general problems of uncertainty-based AL, they were only tested for the variation ratio query function.  This function was chosen, because it was reported in \cite{Gal2016Thesis} to give the best result. To make the results more generalizable, all the experiments mentioned above were run three times. 

Moreover, to test the assumption of the RECT strategy, we measured whether there was a relation between how the model softmax predictions changed towards the one-hot vector of the actual label and the cosine similarity to the data point that was trained on. The relationship was quantified by means of Kendall's $\tau$ between the ranking of the examples based on which one had the largest change in KL divergence after training on the top example and the ranking of the examples based on cosine similarity to the example being trained on.

\subsubsection{Hyperparameters}
Table \ref{tab:hyperparameters} gives an overview of used hyperparameters.
Model weights were randomly initialized using the various PyTorch initialization defaults for the respective model components. In addition to the randomness of weight initialization, randomness determines dropout choices during training. These two forms of randomness influence model performance. For each system/setting, we averaged results over three repeated runs which were identical except for these random elements. This helps to prevent false conclusions due to performance differences caused by effects of these elements.

Both dropout rate and $l$ (the cosine-similarity threshold used in RECS) were chosen based on a grid search across both datasets. The amount of stochastic forward passes $T$ was based on \citet{ein-dor-etal} and was set to 10 across all experiments.\footnote{Larger values up to 100 were tested, but induced much larger training times without noteworthy performance gains.}
Early stopping was applied on each training phase of the AL loop,
Table \ref{tab:early_stopping} shows the amount of epochs used for each dataset. The model yielding the lowest validation loss across all epochs was used for evaluation and uncertainties computation. Note that in a normal AL setting, validation sets are usually not available due to the labelling effort required and this strategy would be less feasible. 

The Adam algorithm \citep{kingma2015adam} was used for optimization and its learning rate was tuned based on the CLR method \citep{DBLP:journals/corr/Smith15a}. The best performing computationally feasible batch size (128), out of the tried batch sizes (32, 64, 128, 256), was used in all experiments. The betas and $\epsilon$ were set to their default values.
The size of $\mathcal{RP}$ was chosen arbitrarily, determining its optimal choice is left future research.

Finally, dimensionality reduction using PCA was tried to determine whether this would result in better class-separability. For every data point in the full dataset, the classes of the group of ten most similar data points (based on cosine similarity) were determined. By maximizing the average of the number of within-group same-class data points, the used dimensionality was determined.

\noindent
\begin{table}[t!]
\begin{minipage}[ht]{0.48\linewidth}
        \centering
        \caption{The amount of epochs used for early stopping for the different datasets.}
        \begin{tabularx}{\linewidth}{XX}
             Dataset & \# Epochs \\
             \hline
             SST & 15\\
             KvK & 25\\
             \hline
        \end{tabularx}
        \label{tab:early_stopping}
\end{minipage}
\hspace{0.04\linewidth}
\begin{minipage}[ht]{0.48\linewidth}
        \centering
        \caption{Hyperparameters values}
        \begin{tabularx}{\linewidth}{XX}
             Parameter & Value \\
             \hline
             Dropout rate & 0.2 \\
             $T$ & 10\\
             $l$ & 0\\
             $\beta_1, \beta_2$ & 0.9, 0.999\\
             $\epsilon$ & 1 * 10$^{-8}$\\
             Learning rate & 2 * 10$^{-5}$\\
             Batch size & 128 \\
             $\mathcal{RP}$ size & 1.5*$q$\\
             Embedding dim. & 768\\
             \hline
        \end{tabularx}
        \label{tab:hyperparameters}
\end{minipage}
\end{table}

\section{Results}\label{sec:Results}
This section will go onto visualize and describe the achieved results for all three experiments described in Section \ref{sssec:experiments}. Note that for all figures, the results were averaged over three runs with the error bars showing one standard deviation. Furthermore, all deficiencies were rounded to two decimal places. For deficiency values  $<$1 (improvements over the reference strategy), we show the smallest value in the comparison in bold. 
For the sake of readability and to keep graph points aligned, in the graphs 
for query-pool sizes of 0.5\% and 1\% the points shown are respectively those at every 10th and 5th and interval.

\subsection{Active Learning}\label{ssec:results_AL}
\begin{figure}[ht!]
    \subfloat[\label{fig:kvk_random_acc}]{%
      \includegraphics[width=0.48\linewidth]{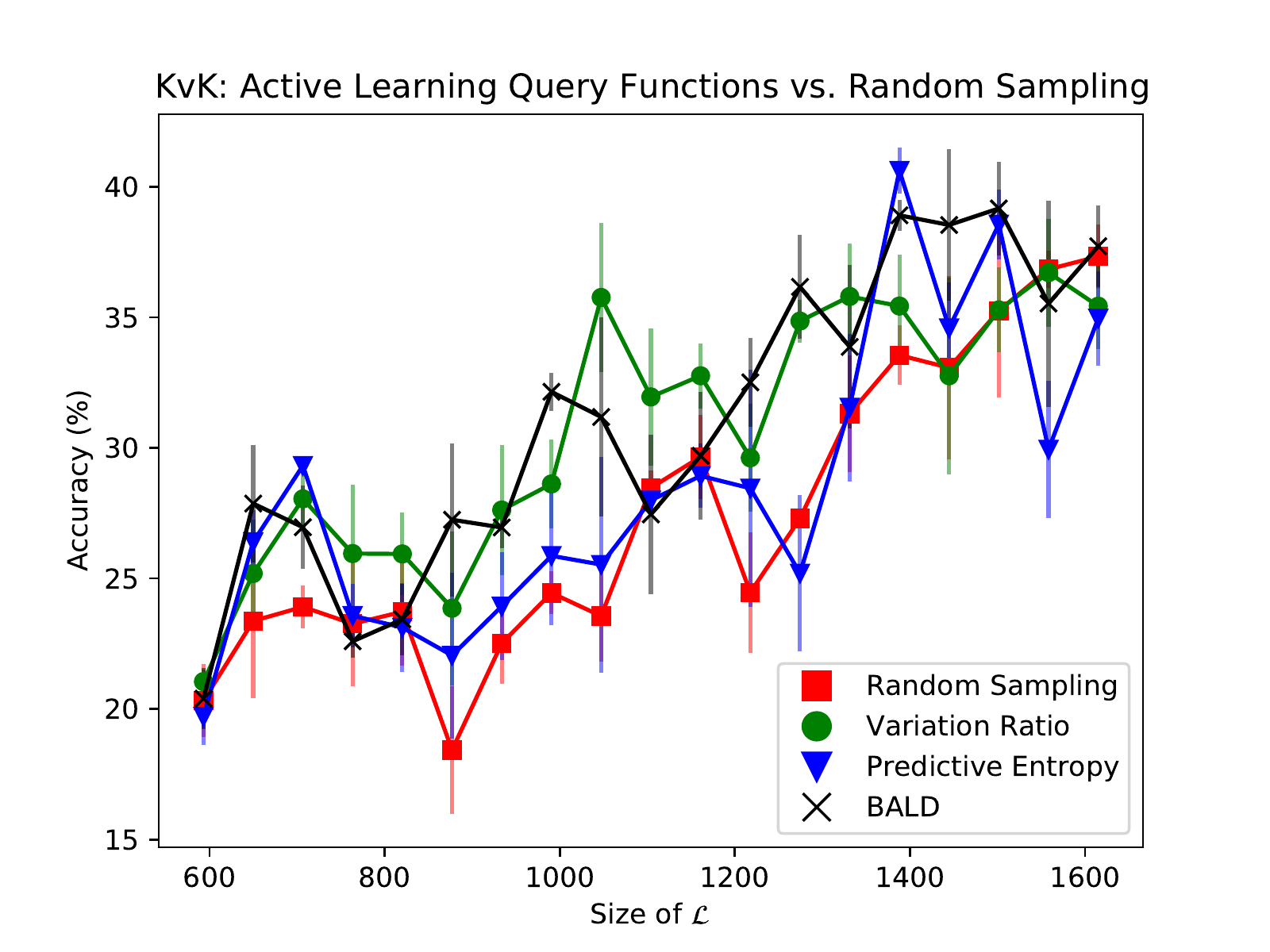} 
    }\hfil
        \subfloat[\label{fig:results_active_learning_sst}]{%
      \includegraphics[width=0.48\linewidth]{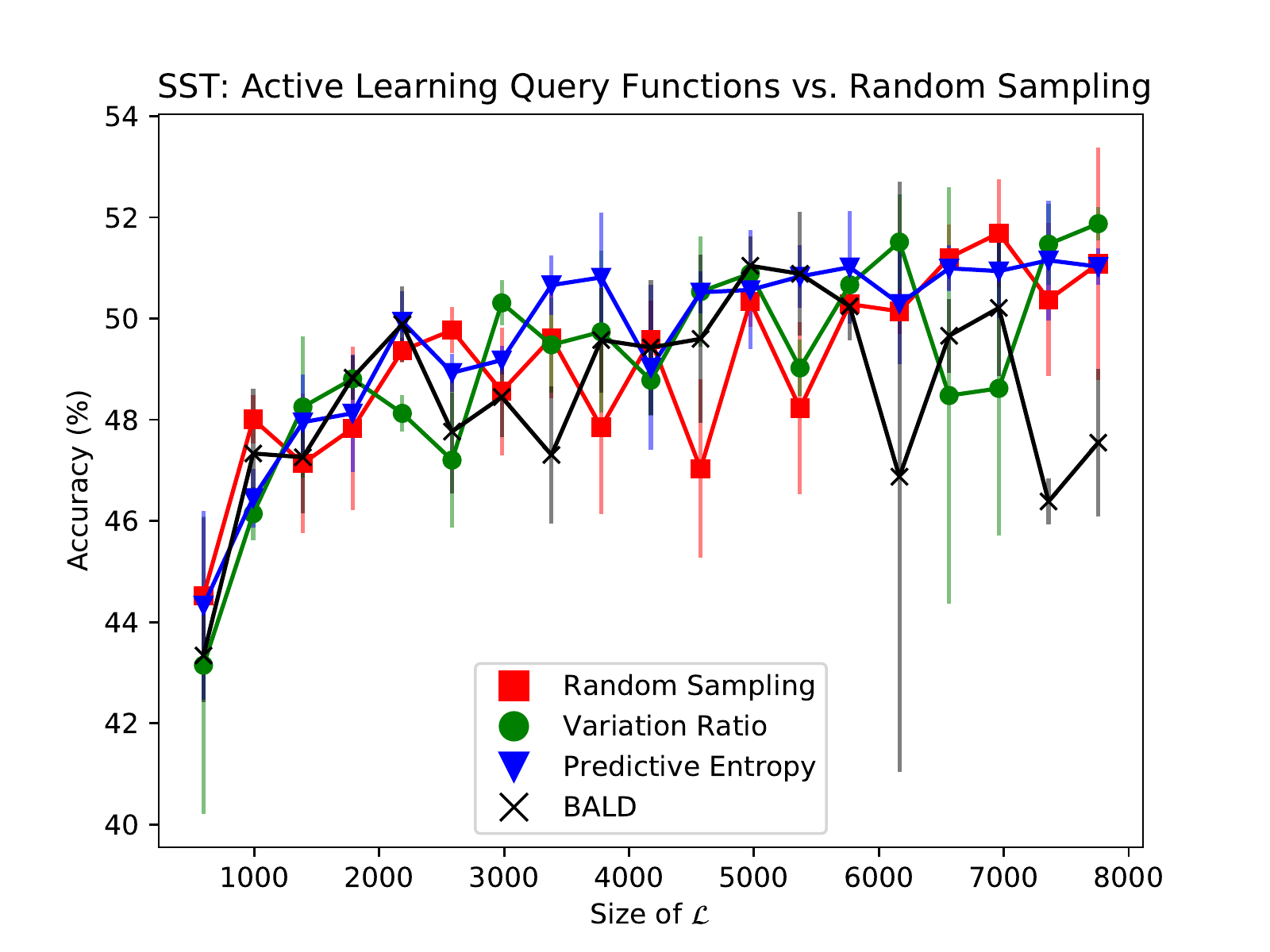}     
    }
    \caption{The achieved test accuracy on the KvK dataset (a) and on the SST dataset (b) by random sampling and the uncertainty-based query functions.} 
\end{figure}

Figure \ref{fig:kvk_random_acc} shows how the query functions performed on the KvK dataset. All query functions outperform random sampling when the labeled dataset is less than 200 examples large. After this, in particular BALD and variation ratio continue to mostly outperform random sampling until near the maximum labeled data size. Notably, many of the performance differences are larger than one standard deviation.

Figure \ref{fig:results_active_learning_sst} shows how random sampling and the implemented query functions performed on the SST dataset. On this dataset the results for the random sampling baseline and the other systems is much smaller, and  there does not seem to be a clear winner.

Finally, the deficiencies shown in Table \ref{tab:deficiency_queryfunctions} show a positive result ($<$ 1) for all query functions except for predictive entropy for the SST dataset. Matching the graphs, the performance gains as measured by the deficiency scores are overall more substantial on the KvK dataset. BALD has the lowest deficiency for both datasets.

\begin{table}[ht]
\centering
    \caption{The deficiencies (Eq. \ref{eq:deficiency}) of the uncertainty-based query functions. Random sampling was the reference strategy.}
    \begin{tabularx}{\linewidth}{XXXX}
            Dataset & VR & PE & BALD \\
            \hline
            SST & 0.95 & 1.01 & \textbf{0.89} \\
            KvK & 0.67 & 0.9 & \textbf{0.64} \\
            \hline
    \end{tabularx}
    \label{tab:deficiency_queryfunctions}
\end{table}
\subsection{Query-pool Size}
\begin{figure}[ht!]
    \subfloat[\label{fig:kvk_scaling_acc}]{%
      \includegraphics[width=0.48\linewidth]{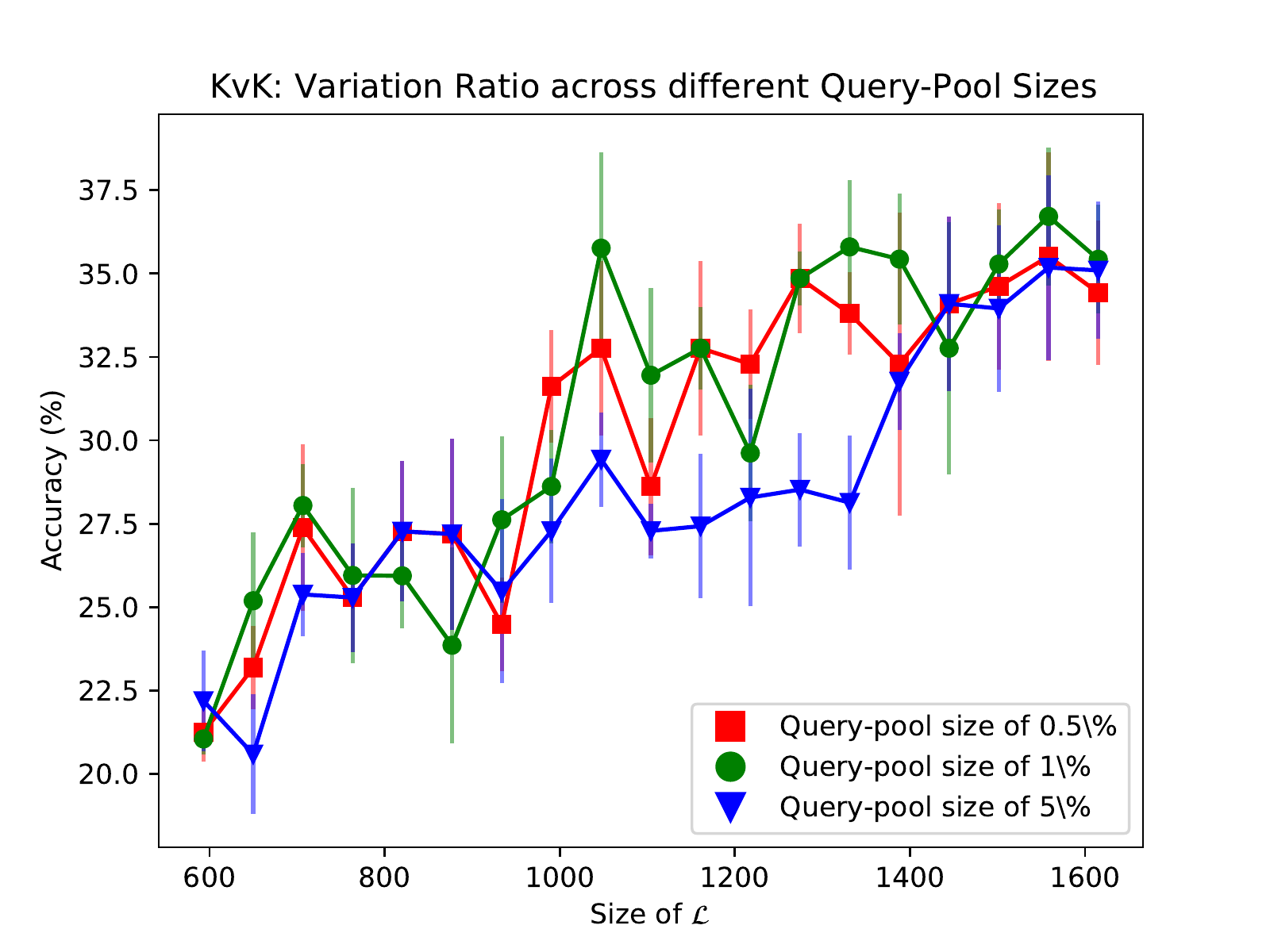} 
     }
    \subfloat[\label{fig:sst_scaling}]{%
      \includegraphics[width=0.48\linewidth]{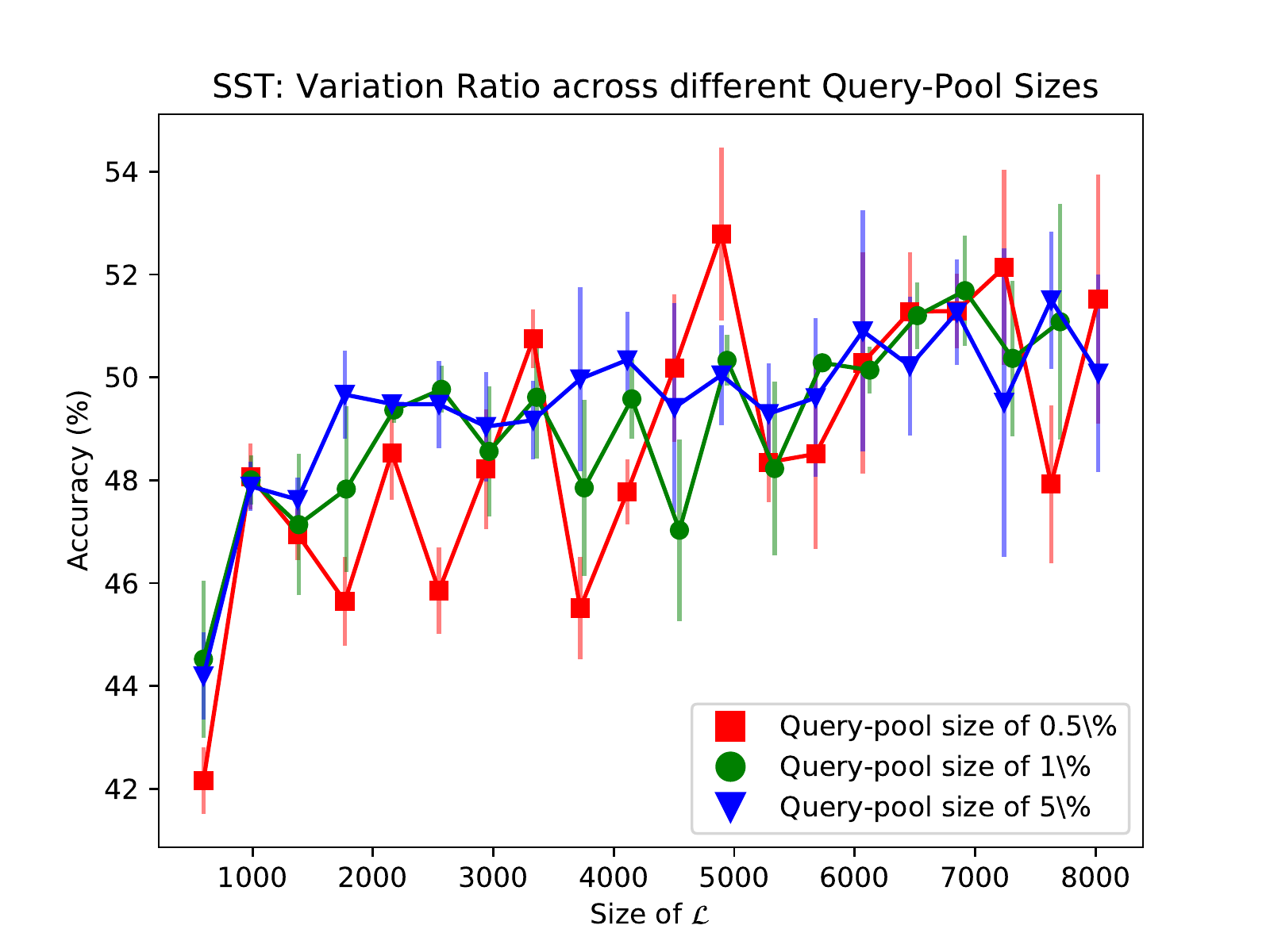} 
     }
        \caption{The achieved test accuracy on the KvK dataset (a) and the SST dataset (b) by using the variation ratio query function with different $q$.}
\end{figure}

Figure \ref{fig:kvk_scaling_acc} shows the performance of variation ratio across different $q$ when used on the KvK dataset.
In the middle range of the graph, variation ratio with a $q$ of 5\% has a worse performance than the other $q$. The $q$ of 0.5\% and 1\% achieve similar performance with the accuracy scores always staying within one standard deviation of each other.

Figure \ref{fig:sst_scaling} shows the performance of the different $q$ on the SST dataset. The performance of variation ratio with a $q$ of 0.5\% fluctuates more when compared to the other $q$.
Moreover, it results in an overall worse performance when compared to the other sizes. The $q$ of 5\% shows to have the best and most consistent performance over the whole learning curve in terms accuracy. However, the $q$ of 0.5\% manages to outperform the other $q$ at about 5000 labeled examples. 

The deficiencies for the different $q$ across both datasets are shown in Table \ref{tab:deficiency_scaling}. For the SST dataset, the $q$ of 5\% had a lower deficiency across the learning curve whereas the $q$ of 0.5\% shows a relatively high deficiency. For the KvK dataset however, we see that the $q$ of 5\% has a relatively high deficiency when compared to the similarly performing $q$ of 0.5\% and 1\%. 
\begin{table}[ht]
     \caption{The achieved deficiencies (Eq. \ref{eq:deficiency}) by the different $q$ for the different datasets. A $q$ of 1\% was the reference strategy.}
        \begin{tabularx}{\linewidth}{XXX}
             Dataset & 0.5\% & 5\% \\
             \hline
             SST & 1.65 & \textbf{0.62}  \\
             KvK & \textbf{0.91} & 1.33\\
             \hline
        \end{tabularx}
    \label{tab:deficiency_scaling}
\end{table} 
\subsection{Heuristics}
\begin{figure}[ht]
  \subfloat[\label{fig:kvk_heuristics_accuracy}]{%
      \includegraphics[width=0.48\linewidth]{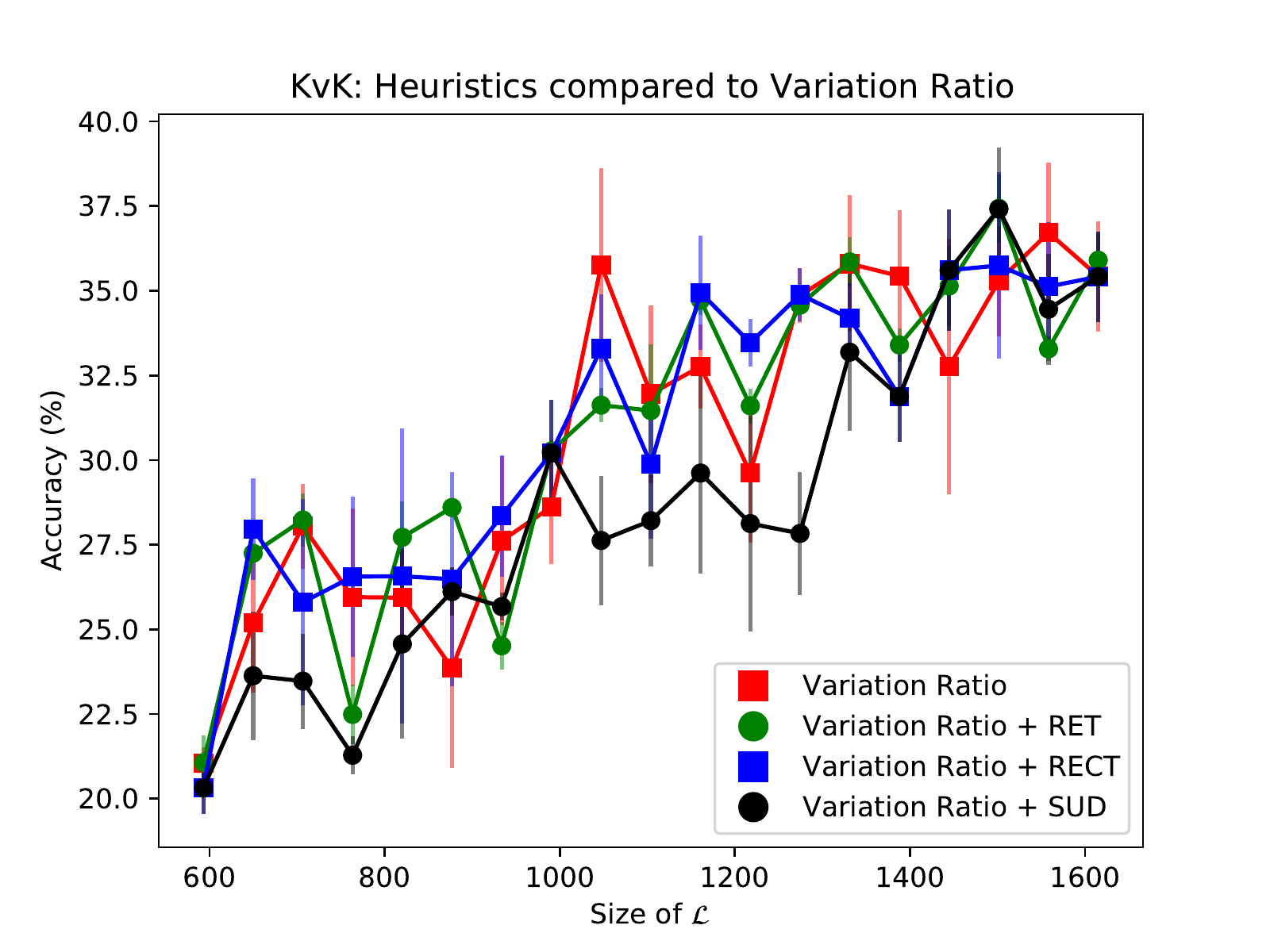} 
   }
   \subfloat[\label{fig:heuristics_sst}]{%
      \includegraphics[width=0.48\linewidth]{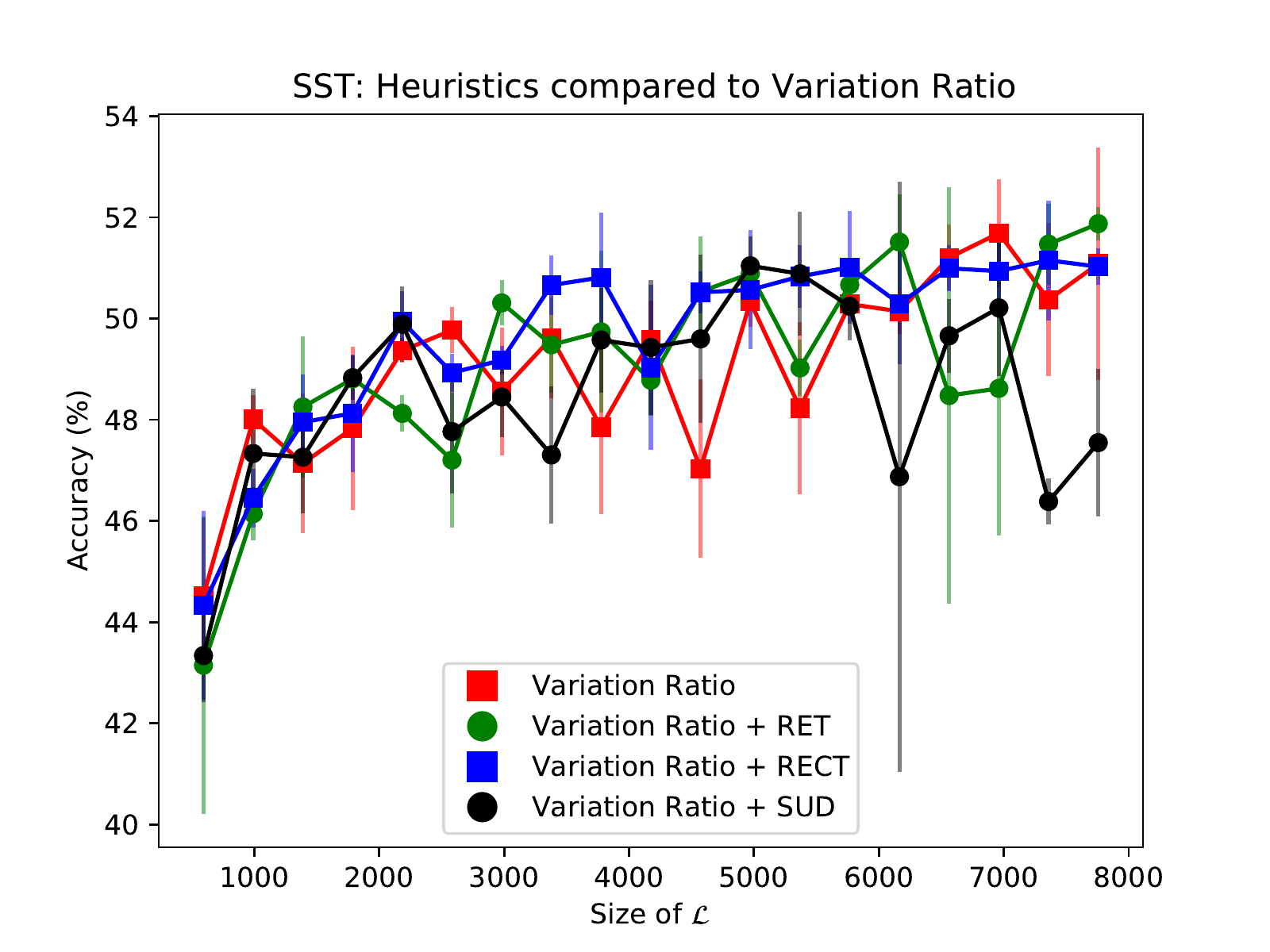} 
   }
    \caption{The achieved test accuracy on the KvK dataset (a) 
    and on the SST dataset (b) by the different heuristics.}
\end{figure}

Figure \ref{fig:kvk_heuristics_accuracy} shows the performance of using variation ratio with heuristics together with the performance of solely using variation ratio on the KvK dataset (also shown in Figure \ref{fig:results_active_learning_sst}). 
Both RET and RECT show no clear improvement over solely using variation ratio. The same can be gathered from the results of the SST dataset shown in Figure \ref{fig:heuristics_sst} as their accuracy scores stay within one standard deviation for the entire learning curve. 
Moreover, Table \ref{tab:ranking_experiment} shows that the average Kendall's $\tau$ is around 0 with a relatively large standard deviation; indicating that there is no relationship between the compared rankings.

Lastly, SUD shows an overall worse performance for both the SST and KvK datasets. 
The deficiencies shown in Table \ref{tab:deficiency_heuristics} also show high values for SUD across both datasets.

\begin{table}[ht]
 \begin{minipage}[ht]{0.48\linewidth}   
    \centering
    \caption{The mean and the 1 SD range of Kendall's $\tau$ from the described ranking experiment across the two datasets (rounded to two decimal places).}
        \begin{tabularx}{\linewidth}{XXX}
            Dataset & Mean & $\sigma$  \\
            \hline
            SST & 0.14 & 0.33 \\
            KvK & 0.02 & 0.47 \\
            \hline
        \end{tabularx}
        \label{tab:ranking_experiment}
\end{minipage}
\hspace{0.2cm}
\begin{minipage}[ht]{0.48\linewidth}
    \centering
       \caption{The achieved deficiencies by the different heuristics. 
       Variation ratio was the reference strategy.}
        \begin{tabularx}{\linewidth}{XXXX}
            Dataset & RET & RECT & SUD \\
            \hline
            SST & 1.02 & 1.05 & 1.23\\
            KvK & 0.98 & \textbf{0.96} & 1.33\\
            \hline
            \end{tabularx}
        \label{tab:deficiency_heuristics}
\end{minipage}
\end{table}

\section{Discussion}\label{sec:Discussion}
This research investigated whether AL could be used to reduce labeling effort while at the same time maintaining similar performance to a model trained on a full dataset. To achieve this, the performance and scalability of different AL query-strategies was tested for the state-of-the-art NLP model: BERT.

\subsubsection{Conclusions}
The results showed that uncertainty-based AL can provide improved performance over random sampling for cut-down datasets. This difference was not consistent throughout the whole training curve: at specific points AL outperformed random sampling and at others at it achieved similar performance. BALD was the query function with the overall best performance. This could be the case due to the fact that it is the only query function used which measures model uncertainty. The found results differs from what was found in \cite{Gal2016Thesis,GalIslamGhahramani2017}, where variation ratio achieved the best overall performance.

Unfortunately, the results found for the KvK dataset show that the found improvement can diminish as query-pool sizes get larger, which corresponds to what was theorized hypothesized in Section \ref{ssec:heuristics}.

Moreover, the two proposed heuristics aimed at improving scalability did not help in improving performance for either dataset and the heuristic aimed at avoiding outliers even resulted in worse performance. This was surprising due to the favorable results found in \cite{zhu-etal-2008-active}, albeit that they only tested it for training sets of up to 150 examples.

An unexpected result was found in that the assumption that semantically similar data conveyed the same type of information did not hold according to the conducted ranking experiment. A possible explanation for this could be that the texts were not mapped to embeddings in a way in which semantically similar data was close enough to each other. Another curious finding was that for the SST dataset, the smallest $q$ resulted in the worse performance, especially at the beginning of the learning curve. This is counter-intuitive due to the fact that performance seems to suffer from more frequent uncertainty estimates. A potential justification for this could be that updating too frequently at the beginning of the learning curve results in the model not being able to train enough on high frequency classes. This could result in the model focusing too much on the long tail of the class distribution due to the fact that it is more uncertain about texts with low frequency classes at the start of the learning curve. Further research is needed to build a better understanding of this. Conversely, given that AL was shown to have little influence on the achieved accuracy and that most of the differences between the different $q$ are within one standard deviation, one could argue that that the size of $q$ did have an influence on the results whatsoever and that we thus cannot conclude anything from the found results.

From the above, we conclude that uncertainty-based AL with BERT$_{base}$ can be used to decrease labeling effort. This supports what was concluded by \cite{BERTActiveLearning2020}. 
    
When looking at the bigger picture, we showed that AL can still provide an improvement in performance over random sampling for large datasets. 
The improvement of performance of AL with BERT is however limited when compared to what it achieved for older NLP models \cite{zhu-etal-2008-active,Tang02activelearning,Mccallum2001} and even more so when compared to image classifiers \citep{houlsby2011bayesian,drost2020,GalIslamGhahramani2017}. Performance did show to increase more when used on the KvK dataset. A possible explanation for this is its smaller size. BERT is pretrained on a large amount of data and only needs fine-tuning for achieving good performance on a specific task. Transfer learning models  \citep{gupta-etal-2020-effective} like BERT have the ability to perform well on new tasks with just a limited amount of data. The power of this few-shot learning also became apparent on a dataset which we decided not to use. Here, BERT was able to get a low validation error on the seed alone, while at the same time having a training accuracy of 100\%.

An additional explanation can be found in the nature of the two tasks and their examples. The SST dataset belongs to a sentiment analysis task, with sentiment scores in the range 0--1. These were binned into spans of 0.2 to get a five-class classification task. Furthermore, bag-of-words (BOW) models such as Naive Bayes were shown to perform relatively really well on this task, because specific individual words provide substantial information about the class. As a consequence, each example is actually \emph{compound}: it indirectly provides information about not just that example but about the sentiment contributions of all the words in that example as well. In contrast, the KvK dataset provides is a real classification task as opposed to a regression task converted to classification task, with 15 distinct classes. A subset of words in each example can be expected to be informative for the class label, as opposed to words giving nearly independent contributions as is the case in sentiment analysis.

A limitation of the research was that, due to computational constraints, only the first sentence of texts was used. There were data points where the first sentence did not contain any clear indication of its label. Take for example the following description from the KvK dataset:
\begin{center}
    \textit{"\textbf{Hi, I'm Barbara Goudsmit.} Welcome to my woven world! I am a passionate hand weaver from the Netherlands who loves creating patterns and bringing them to live on my 8-shaft loom."}
\end{center}

This type of data could have resulted in the network learning suboptimal mappings, which could in turn have had an influence on the performance of AL.

\subsubsection{Future Research}
This work focused on classification tasks. A future direction could be to investigate the influence of AL on BERT's performance in the context of regression tasks and also to examine how the proposed heuristics perform there. 
Moreover, more recent BERT variants, like for instance RoBERTa \citep{Roberta}, could be tested to see whether AL still outperforms the random sampling benchmark. 
Furthermore, the used query functions were mostly developed for and used in computer vision. Query functions aimed at text classification or at the fact that BERT is a pretrained model could be further investigated.
Lastly, an important direction for future work remains making AL more scalable by finding ways to preserve performance with larger query-pool sizes. 

\section*{Acknowledgments}
We would like to express our thanks and gratitude to the people at Dialogic (Utrecht) of which Nick Jelicic in particular, for the useful advice on the writing style of the paper and the suggested improvements for the source code.

%
%
%
%

\bibliographystyle{splncs04}
\bibliography{literature}

\newpage

\section*{Appendix}
\renewcommand{\thesubsection}{A.\arabic{subsection}}

\subsection{RET Algorithm Computational Cost Analysis}
\label{appendix:RET-algorithm-cost-analysis}

The number of forward passes required by the RET algorithm depends on two factors:

\begin{enumerate}
 \item \emph{Basic passes}: The forward passes required by the ``normal'' computation of uncertainty at the beginning of the computation for every query-pool.
 \item \emph{RP passes}: The forward passed required for intermediate updates, using the redundancy pool $RP$.
\end{enumerate}

In this analysis we will assume that the size of the redundancy pool 
$|\mathcal{RP}|$ is chosen as a factor $f > 1$ of the size of the 
query-pool $q$. A reasonable assumption, considering 
that making $|\mathcal{RP}|$ larger than needed incurs unnecessary computational cost, whereas a too small value is expected to diminish the effect of the RET
algorithm.
We furthermore notice that given this assumption, and assuming a fixed total number of examples to label, there are two factors influencing the required amount of \emph{RP passes}: 

\begin{itemize}
 \item Linearly increasing the query-pool size and coupled redundancy pool size 
 causes a quadratic increase in the number of required forward passes per query pool round.
 \item At the same time, a linearly increased query-pool size also induces a corresponding linear decrease in the number of required query-pool rounds.
\end{itemize}

We will see that these two factors will cause a  net linear contribution to the number of \emph{RP passes} starts causing a net increase of total passes once the query-size comes above a certain value. Looking at (1) more precisely, the amount of passes over  $\mathcal{RP}$ that needs to be performed per query-pool round can be computed as an \emph{arithmetic progression}:

\begin{align}
|\mathcal{RP}| + (|\mathcal{RP}| - 1) + (|\mathcal{RP}| - 2) + \ldots +
(|\mathcal{RP} - q) = \\
\frac{1}{2} \times (q + 1) \times (|\mathcal{RP}| + |\mathcal{RP}| - q ) = \\
\frac{1}{2} \times (q + 1) \times ((2f - 1) \times q) = \\
\frac{1}{2} \times (q + 1) \times f' \times q) = \\
\frac{1}{2} \times f' \times (q^2  +  q)) 
\end{align}

Let's assume we use $f = 1.5$ (as also used in our experiments), and 
consequently, $f' = 2f - 1 = 2$.
The number of forward passes over $\mathcal{RP}$ then becomes exactly
$q^2  +  q$.

The complexity can then be expressed by the following formula:

\begin{equation}
 T \times \ceil{\frac{\textrm{\#Samples}}{q}} \times (|\textrm{data}| + q^2  +  q )
\end{equation}

This can be approximately rewritten as:

\begin{align}
 T \times \textrm{\#Samples} \times (\frac{|\textrm{data}|}{q} + \frac{q^2  +  q}{\textrm{query-pool}}) = \\
  T \times \textrm{\#Samples} \times (\frac{|\textrm{data}|}{q} + q + 1)
\end{align}

Note that the second term $\textrm{query-pool-size} + 1$ only starts dominating the number of forward passes in this formula as soon as: 
$$q + 1 \approx q > \frac{|\textrm{data}|}{q} $$ 
This is the case when 
$$q  > \sqrt(|\textrm{data}|)$$ 
Until then, the computational gains of less \emph{basic passes}
outweighs the cost of more \emph{RP passes}.
In practice though, this may happen fairly quickly. For example, assuming we have a data size of $10000$ examples, and we use as mentioned 
$q = 1.5 |\times \mathcal{RP}|$, then as soon as $q \geq 100$ the increased computation of the \emph{RP passes} starts dominating the gains made by less \emph{basic passes} when further increasing the query-pool size, and the net effect is that the total amount of computation increases.

In summary, for the RET algorithm, \emph{RP passes} contribute to the total amount of forward passes. Furthermore, this contribution increases linearly with
redundancy-pool size and coupled query-pool size, and starts to dominate the total amount of forward passes once 
$\textrm{redundancy-pool-size}  > \sqrt{\textrm{data-size}}$. 
This limits its use for decreasing computation by increasing the query-pool size.

\subsection{Algorithms}\label{ap:appendix_algorithm}
\begin{minipage}[t]{\textwidth}
    \begin{algorithm}[H]
        \caption{The general AL loop.}
        \label{alg:al-loop}
        \textbf{Input} Labeled dataset $\mathcal{L}=\{(x_i, y_i)\}^n_i$, the unlabeled data $\mathcal{U}=\{(x_i, \emptyset)\}^n_i$ and the untrained classifier $f(x;\theta)$. \\
        \textbf{Output} Fully labeled dataset $\mathcal{L}=\{(x_i, y_i)\}^n_i$ and trained classifier $f(x;\theta)$
        \begin{algorithmic}[1]
        \STATE $n \gets \text{Desired length of $\mathcal{L}$}$ 
        \STATE  $q \gets \text{Query-pool size}$
        \STATE  $Q(x) \gets \text{Query Function}$
        \WHILE{$\mathcal{L} \text{ length} < n$} 
            \STATE \text{Retrain $f(x;\theta)$ on $\mathcal{L}$}
            \STATE \text{Sort $\mathcal{U}$ based on $Q(\mathcal{U})$}
            \STATE \text{Let Oracle assign labels to $\mathcal{U}^q_0$} 
            \STATE \text{Insert $\mathcal{U}^q_0$ into $\mathcal{L}$}
            \STATE \text{Remove $\mathcal{U}^q_0$ from $\mathcal{U}$}
        \ENDWHILE
        \end{algorithmic}
    \end{algorithm}
 \noindent
    \begin{algorithm}[H]
        \caption{The AL loop with MCD.}
        \label{alg:al-loop-MCD}
        \textbf{Input} Labeled dataset $\mathcal{L}=\{(x_i, y_i)\}^n_i$, the unlabeled data $\mathcal{U}=\{(x_i, \emptyset)\}^n_i$ and the untrained classifier $f(x;\theta)$. \\
        \textbf{Output} Fully labeled dataset $\mathcal{L}=\{(x_i, y_i)\}^n_i$ and trained classifier $f(x;\theta)$
        \begin{algorithmic}[1]
            \STATE $n \gets \text{Desired dataset length}$ 
            \STATE  $q \gets \text{Query-pool size}$
            \STATE  $Q(x) \gets \text{Query Function}$
            \textcolor{ForestGreen}{\STATE $T \gets \text{Number of SFP's}$}
            \WHILE{$\mathcal{L} \text{ length} < n$} 
                \STATE \text{Retrain $f(x;\theta)$ on $\mathcal{L}$}
                \textcolor{ForestGreen}{\STATE $P \gets \emptyset$ 
                \FOR{$t = 0,...,T$}
                    \STATE \text{insert  $f(\mathcal{U}; \theta_t)$ into $P$}
                \ENDFOR
                \STATE \text{Sort $\mathcal{U}$ based on $Q(P)$}}
                \STATE \text{Let Oracle assign labels to $\mathcal{U}^q_0$} 
                \STATE \text{Insert $\mathcal{U}^q_0$ into $\mathcal{L}$}
                \STATE \text{Remove $\mathcal{U}^q_0$ from $\mathcal{U}$}
            \ENDWHILE
        \end{algorithmic}
    \end{algorithm}
   \end{minipage}
\clearpage
\noindent
\begin{minipage}[t]{\textwidth}
    \begin{algorithm}[H]
        \caption{The AL loop with Redundancy Elimination by Training (RET).}
        \label{alg:AL-RET}
        \textbf{Input} Labeled dataset $\mathcal{L}=\{(x_i, y_i)\}^n_i$, the unlabeled data $\mathcal{U}=\{(x_i, \emptyset)\}^n_i$ and the untrained classifier $f(x;\theta)$. \\
        \textbf{Output} Fully labeled dataset $\mathcal{L}=\{(x_i, y_i)\}^n_i$ and trained classifier $f(x;\theta)$
        \begin{algorithmic}[1]
        \STATE $n \gets \text{Desired dataset length}$ 
        \STATE $r \gets \text{Redundancy-pool size}$
        \STATE  $q \gets \text{Query-pool size}$
        \STATE $T \gets \text{Number of SFP's}$
        \STATE  $Q(x) \gets \text{Query Function}$

        \WHILE{$\mathcal{L} \text{ length} < n$} 
            \STATE \text{Retrain $f(x;\theta)$ on $\mathcal{L}$}
            \STATE $P \gets \emptyset$ 
            \FOR{$t = 0,...,T$}
                \STATE \text{insert  $f(\mathcal{U}; \theta_t)$ into $P$}
            \ENDFOR
            \STATE \text{Sort $\mathcal{U}$ based on $Q(P)$}
            \textcolor{ForestGreen}{
            \STATE $ U \gets \emptyset$
            \STATE $queried \gets 0$
            \WHILE{$queried < q$}
                \FOR{$t = 0,...,T$}
                    \STATE \text{insert  $f(\mathcal{RP}; \theta_t)$ into $U$}
                \ENDFOR
                \STATE $i \gets argmin(U)$
                \STATE $\text{Let Oracle assign label to }\mathcal{U}_i$
                \STATE \text{Train $f(x;\theta)$ on $\mathcal{U}_i$}
                \STATE $\text{Insert }\mathcal{U}_i \text{ into } \mathcal{L} $
                \STATE $\text{Remove }\mathcal{U}_i \text{ from } \mathcal{U} $
                \STATE $queried \gets queried + 1$
            \ENDWHILE}
        \ENDWHILE
        \end{algorithmic}
    \end{algorithm}
\end{minipage}
\noindent
\clearpage
\noindent
\begin{minipage}[t]{\textwidth}
    \begin{algorithm}[H]
        \caption{The AL loop with Redundancy Elimination by Cosine Similarity (RECS).}
        \label{alg:RECT}
        \textbf{Input} Labeled dataset $\mathcal{L}=\{(x_i, y_i)\}^n_i$, the unlabeled data $\mathcal{U}=\{(x_i, \emptyset)\}^n_i$ and the untrained classifier $f(x;\theta)$. \\
        \textbf{Output} Fully labeled dataset $\mathcal{L}=\{(x_i, y_i)\}^n_i$ and trained classifier $f(x;\theta)$
        \begin{algorithmic}[1]
        \STATE $n \gets \text{Desired dataset length}$ 
        \STATE $u \gets \text{Redundancy-pool size}$
        \STATE  $q \gets \text{Query-pool size}$
        \STATE $l \gets \text{Cosine similarity threshold}$
        \STATE $T \gets \text{Number of SFP's}$
        \STATE  $Q(x) \gets \text{Query Function}$
        \STATE $Cos(x,y) \gets \text{Cosine similarity between x and y}$
        \WHILE{$\mathcal{L} \text{ length} < n$} 
            \STATE \text{Retrain $f(x;\theta)$ on $\mathcal{L}$}
            \STATE $P \gets \emptyset$ 
            \FOR{$t = 0,...,T$}
                \STATE \text{insert  $f(\mathcal{U}; \theta_t)$ into $P$}
            \ENDFOR
            \STATE \text{Sort $\mathcal{U}$ based on $Q(P)$}
            \textcolor{ForestGreen}{\STATE $ U \gets \emptyset$
            \WHILE{$U length < q$}
                \FOR{$i = 0,...,u$}
                    \IF{$Cos(\mathcal{U}_i, U^{U length}_0) < l$}
                        \STATE \text{insert $\mathcal{U}_i$ into $U$}
                    \ENDIF
                \ENDFOR
                \STATE $l \gets l-0.01$
            \ENDWHILE
            \STATE \text{Reset $l$ to initial value}
            \STATE \text{Let Oracle assign labels to $U$}
            \STATE \text{Insert $U$ into $\mathcal{L}$}
            \STATE \text{Remove $U$ from $\mathcal{U}$}
            }
        \ENDWHILE
        \end{algorithmic}
    \end{algorithm}
\end{minipage}
\clearpage
\noindent
\begin{minipage}[t]{\textwidth}
    \begin{algorithm}[H]
        \caption{The AL loop with SUD.}
        \label{alg:al-SUD}
        \textbf{Input} Labeled dataset $\mathcal{L}=\{(x_i, y_i)\}^n_i$, the unlabeled data $\mathcal{U}=\{(x_i, \emptyset)\}^n_i$ and the untrained classifier $f(x;\theta)$. \\
        \textbf{Output} Fully labeled dataset $\mathcal{L}=\{(x_i, y_i)\}^n_i$ and trained classifier $f(x;\theta)$
        \begin{algorithmic}[1]
        \STATE $n \gets \text{Desired dataset length}$ 
        \STATE  $q \gets \text{Query-pool size}$
        \STATE  $k \gets \text{Amount of similar examples to compute density with}$
        \STATE $T \gets \text{Number of SFP's}$
        \STATE  $Q(x) \gets \text{Query Function}$
        \STATE $Cos(x,y) \gets \text{Cosine similarity between x and y}$
        \WHILE{$\mathcal{L} \text{ length} < n$} 
            \STATE \text{Retrain $f(x;\theta)$ on $\mathcal{L}$}
            \STATE $P \gets \emptyset$
            \STATE $E \gets \emptyset$
            \FOR{$t = 0,...,T$}
                \STATE \text{Insert  $f(\mathcal{U}; \theta_t)$ into $P$}
            \ENDFOR
            \textcolor{ForestGreen}{
            \FOR{$example \text{ in } \mathcal{U}$}
                \STATE $similar \gets Sort(Cos(example, U))$
                \STATE $\text{Insert $\frac{sum(similar_0^k))}{k}$} \text{ into E}$
            \ENDFOR
            }
            \STATE \text{Sort $\mathcal{U}$ based on $Q(P\textcolor{ForestGreen}{*E})$}
            \STATE \text{Let Oracle assign labels to $\mathcal{U}^q_0$} 
            \STATE \text{Insert $\mathcal{U}^q_0$ into $\mathcal{L}$}
            \STATE \text{Remove $\mathcal{U}^q_0$ from $\mathcal{U}$}
        \ENDWHILE
        \end{algorithmic}
    \end{algorithm}
\end{minipage}

\end{document}